\documentclass[jair,twoside,11pt,theapa]{article}
\usepackage{jair, theapa, rawfonts}

\usepackage[pdftex]{graphicx}
\graphicspath{{figures/}}
\DeclareGraphicsExtensions{.pdf,.jpeg,.png}

\usepackage{multirow}

\jairheading{1}{2014}{1-15}{11/13}{?}
\ShortHeadings{Sentiment Analysis}
 {Saif M. Mohammad}

\begin{document}

\title{Sentiment Analysis: Automatically Detecting\\ Valence, Emotions, and Other Affectual States from Text}

\author{\name Saif M. Mohammad \email Saif.Mohammad@nrc-cnrc.gc.ca \\
       \addr National Research Council Canada\\ 
 1200 Montreal Rd., Ottawa, ON, Canada}


\maketitle

\begin{abstract}
Recent advances in machine learning have led to computer systems that are human-like in behaviour. Sentiment analysis, the automatic determination of emotions in text, is allowing us to capitalize on substantial previously unattainable opportunities in commerce, public health, government policy, social sciences, and art. Further, analysis of emotions in text, from news to social media posts, is improving our understanding of not just how people convey emotions through language but also how emotions shape our behaviour. 
This article presents a sweeping overview of sentiment analysis research that includes: the origins of the field,
the rich landscape of tasks, challenges, a survey of the methods and resources used, and applications. 
We also discuss discuss how, without careful fore-thought, sentiment analysis has the potential for harmful outcomes. We outline the latest lines of research in pursuit of fairness in sentiment analysis.\\[3pt]
\end{abstract}

\section{Introduction}
\label{introduction}



{\it Sentiment analysis} is an umbrella term for 
the determination of valence, 
emotions, 
and other affectual states from text or speech automatically using computer algorithms.  
Most commonly, it is used to refer to the task of automatically determining the
valence of a piece of text, whether it is positive, negative, or neutral, the star rating
of a product or movie review, or a real-valued score in a 0 to 1 range that indicates
the degree of positivity of a piece of text.
However, more generally, it can refer to determining one's attitude towards a particular
target or topic.  Here, attitude can mean an evaluative judgment, such as positive or
negative, or an emotional or affectual attitude such as frustration, joy,
anger, sadness, excitement, and so on. Sentiment analysis can also refer to the task of determining
one's emotional state from their utterances (irrespective of whether the text is expressing
an attitude towards an outside entity). The name \textit{sentiment analysis} is a legacy of early work  that focused heavily on appraisals in customer reviews. Since its growth to encompass emotions and feelings, some now refer to the field more broadly as \textit{emotion analysis}.  

\begin{figure}[t]
\centering
 \includegraphics[width=\textwidth]{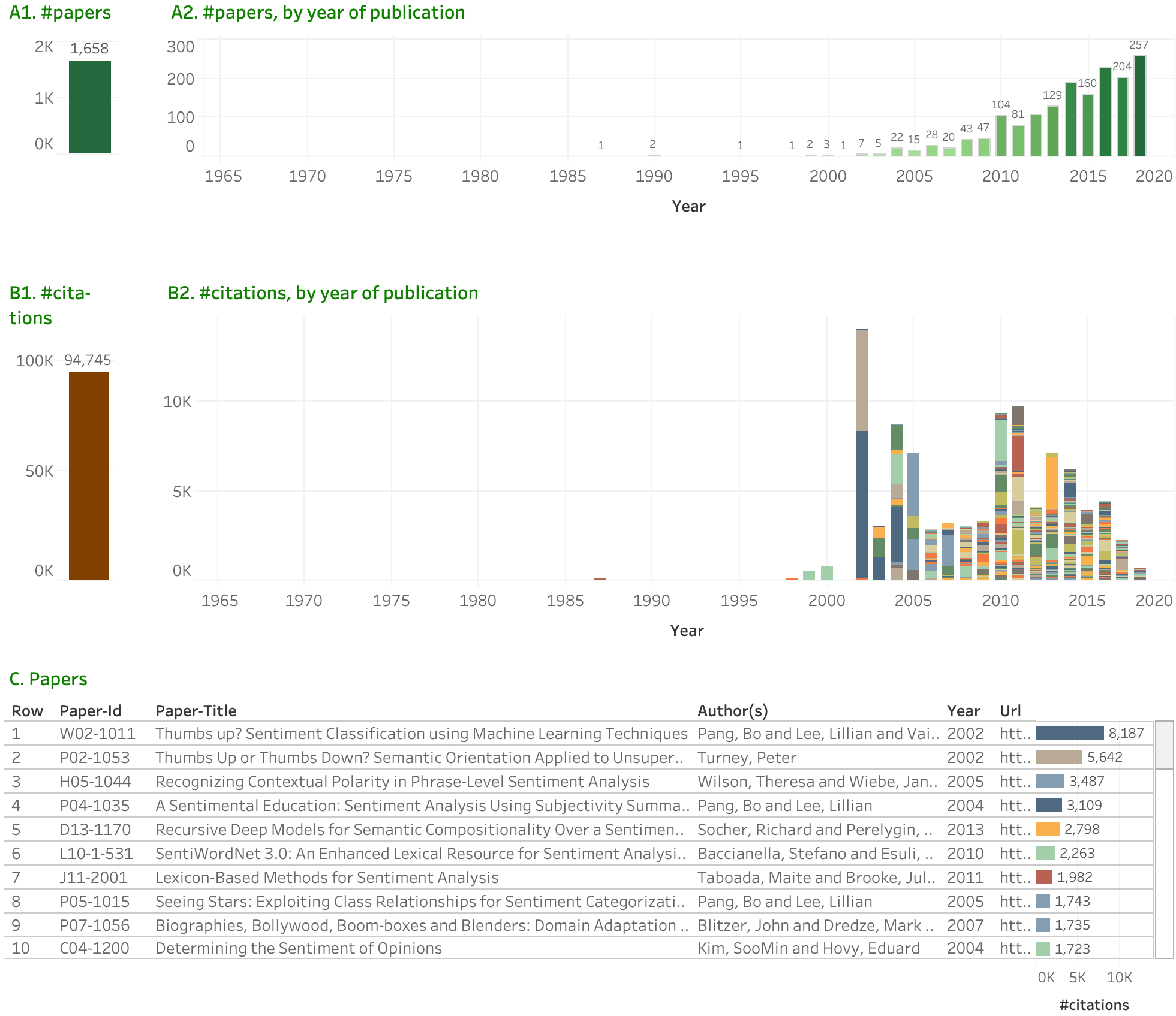}
 \vspace*{3mm}
\caption{A screenshot of the NLP Scholar dashboard when the search box is used to show only those papers that have a sentiment analysis associated term in the title. A1 shows the number of number of papers, and A2 shows the number of papers by year of publication. B1 shows the number of citations for the set (as of June 2019) and B2 shows the citations by year of publication. For a given year, the bar is partitioned into segments corresponding to individual papers. Each segment (paper) has a height that is proportional to the number of citations it has received and assigned a colour at random. C shows the list of papers ordered by number of citations. Hovering over a paper in the interactive visualization shows metadata associated with the paper.}
\label{fig:SAinAA}
\end{figure}

Sentiment analysis, 
as a field of research, arose at the turn of the century with the publication of some highly influential Natural Language Processing (NLP) research.\footnote{See this series of blog posts for a visual analysis of NLP papers, conferences, and journals from 1965 to 2019: https://medium.com/@nlpscholar/state-of-nlp-cbf768492f90. Note  that conference papers in NLP (as in Computer Science) are fully peer-reviewed and published in the proceedings. They are also the predominant mode of publication; far outnumbering journal publications. The top NLP conferences are markedly competitive, with roughly one in four or five submissions being accepted.} 
This initial work was on determining the valence (or sentiment) in customer reviews 
\cite{turney-2002-thumbs,pang-etal-2002-thumbs} 
and on
separating affectual or subjective text from more factual and non-affective text \cite{wiebe-etal-1999-development,riloff-wiebe-2003-learning}.
The platform for this work was set in the 1990's with the growing availability of large amounts of digitally accessible text
as well as, of course, earlier seminal work at the intersection of emotions and language in
psychology, psycolinguistics, cognitive psychology, and behavioural science \cite{Osgood57,russell1980circumplex,oatley1987towards}. 
See Figure \ref{fig:SAinAA} (a) for a timeline graph of the number of papers in the ACL Anthology with sentiment and associated terms in the title. 
(The ACL Anthology is a repository of public domain, free to access, articles on NLP with more than 50K articles published since 1965.\footnote{https://www.aclweb.org/anthology/})
Figure \ref{fig:SAinAA} (b) shows a timeline graph where each paper is represented as a segment whose height is proportional to the number of citations it has received as of June 2019.\footnote{Terms used for the query in the visualization: \textit{sentiment, valence, emotion, emotions, affect, polarity, subjective, subjectivity,} and \textit{stance}. NLP Scholar: http://saifmohammad.com/WebPages/nlpscholar.html}
Observe the large citations impact of the 2002 papers by \citeA{turney-2002-thumbs} and \citeA{pang-etal-2002-thumbs} and subsequent papers published in the early to mid 2000s. Since then, the number of sentiment analysis papers published every year has steadily increased.

Several notable works on the relationship between language and sentiment were carried out much earlier, though. 
\citeA{Osgood57} asked human participants to rate words along dimensions
of opposites such as {\it heavy--light, good--bad, strong--weak,} etc.
Their seminal work on these judgments showed that the most prominent dimension of connotative word meaning
is evaluation (aka valence) ({\it good--bad, happy--sad}), followed by potency ({\it strong--weak, dominant--submissive}) and activity ({\it active--passive, energetic--sluggish}).
\citeA{russell1980circumplex} 
showed through similar analyses of emotion words that the three primary 
dimensions of emotions are valence (pleasure--displeasure, happy--sad), 
arousal (active--passive), and dominance (dominant--submissive). 
\citeA{OrtonyCC88} 
argued that all emotions are valenced, that is, emotions are either positive or negative, but never neutral \cite{OrtonyCC88}. While 
instantiations of some emotions tend to be associated with exactly one valence (for example, joy is 
always associated with positive valence), instantiations of other emotions may be associated with differing valence
(for example, some instances of surprise are associated with positive valence, while some others are associated
with negative valence). 
Even though Osgood, Russell, and Ortony chose slightly different words as representative of the primary dimensions,
e.g. evaluation vs.\@ valence, potency vs.\@ dominance, and activity vs.\@ arousal,
their experiments essentially led to the same primary dimensions of emotion or connotative meaning.
For the purposes of this chapter, we will use the terms coined by \citeA{russell1980circumplex}:
valence (V), arousal (A), and dominance (D).

In contrast to the VAD model, Paul Ekman and others developed theories on how some emotions are more basic, more important, or more universal, than others \cite{Ekman92,ekman2003unmasking,Plutchik80,plutchik1991emotions}.
 These emotions, such as joy, sadness, fear, anger, etc.\@, are argued to have ties to universal facial expressions and physiological processes such as increased heart rate and perspiration.
\citeA{Ekman92}, \citeA{Plutchik80}, \citeA{Parrot01}, and \citeA{frijda1988laws} 
 and others proposed different sets of basic emotions.
However, these assertions of universal ties between emotions and facial expressions, as well 
as the theory of basic emotions have been challenged and hotly debated in recent work \cite{barrett2006emotions,de2015cultural}. 

 

This chapter presents a comprehensive overview of work on automatically detecting valence, emotions, and other affectual states from \textit{text}.\footnote{See surveys by \citeA{el2011survey} and \citeA{anagnostopoulos2015features} for an
overview of emotion detection in \textit{speech}.  See \citeA{picard2000affective} and
\citeA{alm2008affect} for a broader introduction of giving machines the ability to detect
sentiment and emotions in various modalities such as text, speech, and vision. See articles by \citeA{lawrence2020argument} and \citeA{cabrio2018five} for surveys on the related area of argument mining.} We begin
in Section 2 
by discussing various challenges to sentiment analysis, including: the subtlety with emotions can be conveyed,
the creative use of language, the bottleneck of requiring annotated data, and the lack of para-linguistic context. 
In Section 3, we 
describe the diverse landscape of sentiment analysis problems, including:
detecting sentiment of the writer, reader, and other relevant entities; 
detecting sentiment from words, sentences, and documents; 
detecting stance towards events and entities which may or may not be explicitly mentioned in the text; detecting
sentiment towards aspects of products; and detecting semantic roles of feelings.

Sections 4 delves into approaches to automatically detect emotions from text (especially, sentences and tweets). 
We discuss the broad trends in machine learning that have swept across Natural Language Processing (NLP) and sentiment analysis,
including: transfer learning, deep neural networks, and the representation of words and sentences with dense, low-dimensional, vectors. 
We also identify influential past and present work, along with influential annotated resources for affect prediction.
Section 5 summarizes work on creating emotion lexicons, both manually and automatically.
Notably, we list several influential lexicons pertaining to the basic emotions as well as valence, arousal, and dominance.
Section 6 explores work on determining the impact of sentiment modifiers, such as negations, intensifiers, and modals.

Section 7 discusses some preliminary sentiment analysis work  on sarcasm, metaphor, and other figurative language.
Since much of the research and resource development in sentiment analysis has been on English texts,
sentiment analysis systems in other languages tend to be less accurate. This has ushered
work in leveraging the resources in English for sentiment analysis in the resource-poor languages.
We discuss this work in Section 8.

Section 9 presents prominent areas where sentiment analysis is being applied: from commerce and intelligence gathering to policy making, public, health, and even art. However, it should be noted that these applications have not always yielded beneficial results.
Recent advances in machine learning have meant that computer systems are becoming more human-like in their behaviour. This also means that they perpetuate human biases. Some learned biases may be beneficial for the downstream application. Other biases can be inappropriate and result in negative experiences for some users. Examples include, loan eligibility and crime recidivism systems that negatively assess people belonging to a certain area code (which may disproportionately impact people of certain races) \cite{chouldechova2017fair} and resum\'e sorting systems that believe that men are more qualified to be programmers than women \cite{bolukbasi2016man}. Similarly, sentiment analysis systems can also perpetuate and accentuate inappropriate human biases. We discuss issues of fairness and bias in sentiment analysis in Section 10.

\section{Challenges in Sentiment Analysis}
There are several challenges to automatically detecting sentiment in text:\\[-10pt]

\noindent {\it Complexity and Subtlety of Language Use:}
\begin{itemize}
\item The emotional import of a sentence or utterance is not simply the sum of
emotional associations of its component words. 
Further, emotions are often not stated explicitly. For example, consider:\\[3pt]
\hspace*{10mm} {\it Another Monday, and another week working my tail off.}\\[3pt]
It conveys a sense of frustration without the speaker explicitly saying so. Note also that the sentence
does not include any overtly negative words. \\
Section \ref{sec:autosystems} summarizes various machine learning approaches for classifying sentences and tweets into one of the affect categories.
\item Certain terms such as
negations and modals impact sentiment of the sentence, without themselves having strong sentiment associations.
For example, {\it may be good}, {\it was good}, and {\it was not good} should be interpreted differently by sentiment analysis systems.\\
Section \ref{sec:mods} discusses approaches that explicitly handle
sentiment modifiers such as negations, degree adverbs, and modals.
\item Words when used in different contexts (and different senses) can convey different emotions.
For example, the word {\it hug} in the {\it embrace} sense, as in:\\[3pt]
\hspace*{10mm} {\it Mary hugged her daughter before going to work.}\\[3pt]
\noindent is associated with 
joy and affection. However, in the sentence below:\\[3pt] 
\hspace*{10mm} {\it The pipeline hugged the state border.}\\[3pt]
{\it hug} in the {\it stay close to sense}, and is rather unemotional. Word sense disambiguation remains a
challenge in natural language processing \cite{kilgarriff1997don,navigli2009word}.\\
In Section \ref{sec:lex}, we discuss approaches to create term--sentiment association lexicons,
including some that have separate entries for each sense of a word.
\item Utterances may convey more than one emotion (and to varying degrees). They may convey 
contrastive evaluations of multiple target entities. 
\item Utterances may refer to emotional events without
implicitly or explicitly expressing the feelings of the speaker. \\[-6pt]

\hspace*{-1cm} {\it Use of Creative and Non-Standard Language:}
\item Automatic natural language systems find it difficult to interpret creative uses of language
such as sarcasm, irony, humour, and metaphor. However, these phenomenon are common in language use.\\
Section \ref{sec:figurative} summarizes some preliminary work in this direction.
\item Social media texts are rife with terms not seen in dictionaries, including misspellings ({\it parlament}), 
creatively-spelled words ({\it happeee}), hashtagged words ({\it \#loveumom}),
emoticons, abbreviations ({\it lmao}), etc. Many of these terms convey emotions.\\
Section \ref{sec:autolex} describes work on automatically generating term--sentiment association lexicons
from social media data---methods that capture sentiment associations of not just regular English
terms, but also terms commonly seen in social media.\\[-6pt]

\hspace*{-1cm} {\it Lack of Para-Linguistic Information:}
\item Often we communicate affect through tone, pitch, and emphasis.
However, written text usually does not come with annotations of stress and intonation.
This is compensated to some degree by the use of explicit emphasis markers (for example, Mary used *Jack's* computer) 
and explicit sentiment markers such as emoticons and emoji.
\item We also communicate emotions through facial expressions. In fact there is a lot of work linking
different facial expressions to different emotional states \cite{Ekman92,ekman2003unmasking}.\\[-6pt]

\hspace*{-1cm} {\it Lack of Large Amounts of Labeled Data:}
\item Most machine learning algorithms for sentiment analysis require significant
amounts of training data (example sentences marked with the associated emotions).
However, there are numerous affect categories including hundreds of emotions that humans can perceive and express.
Thus, much of the work in the community has been restricted to a handful of emotions and valence categories.\\
Section \ref{sec:emosystems} summarizes various efforts to create datasets that have sentences labeled with emotions.\\[-6pt]

\hspace*{-1cm} {\it Subjective and Cross-Cultural Differences:}
\item Detecting emotions in text can be difficult even for humans. 
The degree of agreement between annotators is significantly lower when assigning valence or emotions
to instances, as compared to tasks such as identifying part of speech and detecting named entities.
\item There can be significant differences in emotions associated with events and behaviors across different cultures.
For example, {\it dating} and {\it alcohol} may be perceived as significantly more negative in some parts of the world
than in others. 
\item Manual annotations can be significantly influenced by
clarity of directions, difficulty of task, training of the respondents, and even the annotation scheme (multiple choice questions, free text, Likert scales, etc.).\\
Sections \ref{sec:autosystems} and \ref{sec:lex} describe various manually annotated datasets where affect labels are provided for sentences and words, respectively.
They were created either by hand-chosen
expert annotators, known associates and grad students, or by crowdsourcing on the world wide web to hundreds or
thousands of unknown respondents. 
Section \ref{sec:realval} describes an annotation scheme called {\it best--worst scaling (BWS)} 
that has led to more high-quality and consistent sentiment annotations.\footnote{Maximimum Difference Scaling (max-diff) is a slight variant of BWS, and sometimes the terms are used interchangeably.}

\end{itemize}
\noindent 
In the sections ahead we describe approaches that, to some extent, address these issues.
Nonetheless, significant progress still remains to be made.
 

\section{Sentiment Analysis Tasks} 
Even though early work focused extensively on the sentiment in customer reviews, 
sentiment analysis involves a diverse landscape of tasks. Thus, when building or using sentiment analysis
systems, it is important to first determine the precise task that is intended: for example,
whether it is to determine the sentiment of the reader or the writer, whether it is to determine attitude towards a product or one's emotional state, whether it is intended to assess sentences, tweets, or documents, etc. We discuss the landscape of sentiment analysis tasks below.
 

\subsection{Detecting Sentiment of the Writer, Reader, and other Entities}

Sentiment can be associated with any of the following: 1. the speaker (or writer), 2. the
listener (or reader), or 3.  one or more entities mentioned in the utterance.  
Most research in sentiment analysis has focused on detecting the sentiment of the
speaker, and this is often done by analyzing only the utterance.
However, there are several instances where it is unclear whether the sentiment in the utterance
is the same as the sentiment of the speaker. For example, consider:
\begin{quote}
Sarah: {\it The war in Syria has created a refugee crisis.}
\end{quote}
The sentence describes a negative event (millions of people being displaced), but it is
unclear whether to conclude that Sarah (the speaker) 
is personally saddened by the event. It is possible that Sarah is a news reader and merely communicating
information about the war.  Developers of sentiment systems have to decide before hand
whether they wish to assign a negative sentiment or neutral sentiment to the speaker in such cases.
More generally, they have to decide whether the speaker's sentiment will be chosen to be
neutral in absence of overt signifiers of the speaker's  own sentiment, or whether the
speaker's sentiment will be chosen to be the same as the sentiment of events and topics
mentioned in the utterance.

On the other hand, people can react differently to the same utterance, for example, people
on opposite sides of a debate or rival sports fans. Thus modeling listener sentiment
requires modeling listener profiles. This is an area of research not explored much by the
community.  Similarly, there is little work on modeling sentiment of entities mentioned in the
text, for example, given:
\begin{quote}
Drew: {\it Jamie could not stop gushing about the new Game of Thrones episode.}
\end{quote}
\noindent It will be useful to develop automatic systems that can deduce that Jamie (not
Drew) liked the new episode of {\it Game of Thrones} (a TV show).

\subsection{Detecting Sentiment from Different Textual Chunks}
\label{sec:chunks}

Sentiment can be determined at various levels of text: from sentiment associations of words and
phrases; to sentiments of sentences, SMS messages, chat messages, and tweets.
One can also explore sentiment in product reviews, blog posts, whole documents, and even streams of texts
such as tweets mentioning an entity over time.\\[-10pt]  

\noindent {\bf Words:} Some words denotatate valence, i.e., valence is part of their core meaning, for example,
{\it good, bad, terrible, excellent, nice,} and so on. Some other words do not denotate
valence, but have strong positive or negative associations (or connotations). 
For example, \textit{party} and \textit{raise} are associated with positive valence,
whereas \textit{slave} and {\it death} are associated with negative valence. 
Words that are not strongly associated with positive or negative valence are considered
neutral. (The exact boundaries between neutral and positive valence, and between neutral and negative
valence, are somewhat fuzzy. However, for a number of terms, there is high inter-rater agreement on
whether they are positive, neutral, or negative.)
Similarly, some words express emotions as part of their meaning (and thus are also associated with the emotion),
and some words are just associated with emotions. For example,
{\it anger} and {\it rage} denote anger  (and are associated with anger), whereas 
{\it negligence, fight,} and {\it betrayal} do not denote anger, but they are 
 associated with anger.

Sentiment associations of
words and phrases are commonly captured in valence and emotion association lexicons.  A
valence (or polarity) association lexicon may have entries such as these shown below (text
in parenthesis is not part of the entry, but our description of what the entry indicates):
\begin{quote}
	{\it delighted} -- positive ({\it delighted} is usually associated with positive valence)\\
	{\it death} -- negative ({\it death} is usually associated with negative valence)\\
	{\it shout}  -- negative ({\it shout} is usually associated with negative valence)\\
	{\it furniture} -- neutral ({\it furniture} is {\bf not} strongly associated with positive or negative valence)
\end{quote}
\noindent An affect association lexicon has entries for a pre-decided set of emotions (different lexicons
may choose to focus on different sets of emotions). Below are examples of some affect association entries:
\begin{quote}
	{\it delighted} -- joy ({\it delighted} is usually associated with the emotion of joy)\\
	{\it death} -- sadness ({\it death} is usually associated with the emotion of sadness)\\
	{\it shout} -- anger  ({\it shout} is usually associated with the emotion of anger)\\
	{\it furniture} -- none ({\it furniture} is {\bf not} strongly associated with any of the pre-decided set of emotions)
\end{quote}
\noindent A word may be associated with more than one emotion, in which case, it will have
more than one entry in the affect lexicon.  

Sentiment association lexicons can be created either by manual
annotation or through automatic means.  Manually created lexicons tend to have 
a few thousand entries, but automatically generated lexicons can capture valence and
emotion associations for hundreds of thousands unigrams (single word strings) and even for
larger expressions such as bigrams (two-word sequences) and trigrams (three-word
sequences).  Automatically generated lexicons often also include a real-valued score
indicating the strength of association between the word and the affect category.  This
score is the prior estimate of the sentiment association, calculated from previously seen usages of the term.
While sentiment lexicons are often useful in sentence-level sentiment
analysis,
the same terms may convey different sentiments in different contexts.  
The top systems \cite{duppada-etal-2018-seernet,agrawal-suri-2019-nelec,huang-etal-2019-ana,abdou-etal-2018-affecthor,MohammadSemEval2013,Kiritchenko_SemEval2014,Zhu_SemEval2014,tang2014coooolll} in 
recent sentiment-related shared tasks, The SemEval-2018 Affect in Tweets, SemEval-2013 and 2014 Sentiment Analysis in Twitter,  
used large sentiment lexicons \cite{SemEval2018Task1,chatterjee-etal-2019-semeval,SemEval2013task2,SemEval2014task9}.\footnote{https://www.cs.york.ac.uk/semeval-2013/\\http://alt.qcri.org/semeval2014/}
(Some of these tasks also had separate sub-tasks aimed at
identifying sentiment of terms in context.) 
We discuss manually and automatically
created valence and emotion association lexicons in more detail in Section \ref{sec:lex}.\\[-10pt]

\noindent {\bf Sentences:} Sentence-level valence and emotion classification systems assign labels such as positive, negative, or
neutral to whole sentences.  It should be noted that the valence of a sentence is not
simply the sum of the polarities of its constituent words.  Automatic systems learn a
model from labeled training data (instances that are already marked as positive, negative,
or neutral) as well as a large amount of (unlabeled) raw text using 
low-dimensional vector representations of the sentences and constituent words,
as well as more traditional features such as those drawn from word and character ngrams, valence
and emotion association lexicons, and negation lists. 
We discuss the valence (sentiment) and emotion classification systems and the resources they commonly use
 in Sections \ref{sec:valsystems} and \ref{sec:emosystems}, respectively.\\[-10pt]


\noindent {\bf Documents:} Sentiment analysis of documents is often broken down into the sentiment
analysis of the component sentences.  
Thus we do not discuss this topic in much detail here.
However, there is interesting work on using
sentiment analysis to generate text summaries
\cite{ku2006opinion,somprasertsri2010mining} 
and on analyzing patterns of sentiment in social networks in novels and fairy tales
\cite{nalisnick2013character,MohammadY11,davis14}.\\[-10pt] 

\noindent {\bf Document and Twitter Streams:} Sentiment analysis has also been applied to streams of documents and twitter streams where the purpose is usually to detect aggregate trends in emotions over time. This includes work on
determining media portrayal of events by analyzing online news streams \cite{liu2016dynamic}
and predicting stock market trends through the sentiment in financial news articles \cite{schumaker2012evaluating},
 finance-related blogs \cite{o2009topic}, and tweets \cite{smailovic2014stream}.
There is also considerable interest in tracking public opinion over time: \citeA{Thelwall11} tracked sentiment towards 30 events in 2010 including
the Oscars, earth quakes, tsunamis, and celebrity scandals; \citeA{bifet2011detecting} analyzed sentiment in tweets pertaining to the 2010 Toyota crisis; 
\citeA{kim2016topic} tracked sentiment in Ebola-related tweets;
\citeA{vosoughi2018spread} published influential work that showed that false stories elicited responses of fear, disgust, and surprise, whereas true stories elicited responses with anticipation, sadness, joy, and trust;
\citeA{fraser2019we} analyzed emotions in the tweets mentioning a hitch-hiking Canadian robot;
and it is inevitable that soon there will be published work on tracking various aspects of the
global Covid-19 pandemic.

It should be noted that the analyses of streams of data have unique challenges, most notable in the drift of topic of discussion. Thus for example, if one is to track the tweets pertaining to the Covid-19 pandemic, one would have started by polling the Twitter API for tweets with hashtags \#coronavirus and \#wuhanvirus, but will have to update the the search hashtags continually over time to terms such as \#covid19, \#socialdistancing, \#physicaldistancin, \#UKlockdown, and so on.
Yet another, notable challenge in aggregate analysis such as that in determining public emotional state through social media are the biases that impact positing behaviour.
For example, it is well known that people have a tendency to talk more about positive feeling and experiences to show that they are happy and successful \cite{meshi2013nucleus,jordan2011misery}. 
In contrast, when reporting about products there is a bias towards reporting shortcomings and negative experiences \cite{hu2009overcoming}.
It is also argued that there is a bias in favor of report extreme high-arousal emotions because such posts are more likely to spread and go viral \cite{berger2013contagious}.

An offshoot of work on the sentiment analysis of streams of text is work on visualizing emotion in these text streams
\citeA{lu2015visualizing} visualized sentiment in geolocated ebola tweets;
\citeA{gregory2006user} visualized reviews of \textit{The Da Vinci Code}; and
\citeA{fraser2019we} visualized emotions in tweets before and after the death of HitchBot and across countries.
See survey articles by \citeA{boumaiza2015survey} and \citeA{kucher2018state} for further information on sentiment visualization techniques.

\subsection{Detecting Sentiment Towards a Target}

\subsubsection{Detecting Sentiment Towards Aspects of an Entity or Aspect Based Sentiment Analysis (ABSA)}

A review of a product or service can express sentiment towards various aspects.  For
example, a restaurant review can praise the food served, but express anger towards
the quality of service. There is a large body of work on detecting aspects of
products and also sentiment towards these aspects
\cite{schouten2015survey,Popescu05,Su06,Xu12,Qadir09,Zhang10,Kessler2009}.  In 2014, a shared task
was organized for detecting aspect sentiment in restaurant and laptop reviews
\cite{SemEval2014}.  The best performing systems had a strong sentence-level sentiment
analysis system to which they added localization features so that more weight was given to
sentiment features close to the mention of the aspect.  This task was repeated in 2015 and 2016.
It will be useful to develop aspect-based sentiment systems for other domains such as
blogs and news articles as well.  (See proceedings of the ABSA tasks in SemEval-2014, 2015, and 2016 for details
about participating aspect sentiment systems.\footnote{http://alt.qcri.org/semeval2014/\\http://alt.qcri.org/semeval2015/})
As with various task in NLP, over the last year, a number of distant learning based approaches for ABSA
have been proposed, for example \citeA{li2019exploiting} develop a BERT-based benchmark for the task.
See surveys by \citeA{do2019deep,laskari2016aspect,schouten2015survey,vohra2013applications} for further information on aspect based sentiment analysis.

\subsubsection{Detecting Stance}

Stance detection is the task of automatically determining from text whether the author of
the text is in favor or against a proposition or target. 
Early work in stance detection, focused on congressional debates \cite{thomas2006get} or debates in online forums
\cite{Somasundaran2009,murakami-raymond:2010:POSTERS,anand2011cats,walker2012stance,hasan2013stance,sridhar2014collective}. However, more recently, there has been a spurt of work on social media texts.
The first shared task on detecting stance in tweets was organized in 2016 \citeA{mohammad2016semeval,mohammad2017stance} .
They framed the task as follows: Given a tweet text and a pre-determined target proposition,
state whether the tweeter is likely in favor of the proposition, against the proposition, or whether
neither inference is likely. For example,
given the following target and text pair:
\vspace*{-1mm}
\begin{quote}
 \hspace*{0mm} Target: {\it Pro-choice movement OR women have the right to abortion}\\
 \hspace*{0mm} Text: \hspace*{2mm}     {\it A foetus has rights too!}
\end{quote}
\vspace*{-1mm}
\noindent Automatic systems have to deduce the likely stance of the tweeter towards the target.
Humans can deduce from the text that the speaker is against the proposition.
However, this is a challenging task for computers. To successfully detect stance,
automatic systems often have to identify relevant bits of information that may not be
present in the focus text. For example, that if one is actively supporting foetus rights,
then he or she is likely against the right to abortion. Automatic systems can obtain such
information from large amounts of existing unlabeled text about the target.\footnote{Here `unlabeled' 
refers to text that is not labeled for stance.}

Stance detection is related to sentiment analysis, but the two have significant
differences. In sentiment analysis, systems determine whether a piece of text is positive,
negative, or neutral. However, in stance detection, systems are to determine favorability
towards a given target -- and the target may not be explicitly mentioned in the text. For
example, consider the target--text pair below:
\vspace*{-1mm}
\begin{quote}
\hspace*{0mm} Target: {\it Barack Obama}\\
\hspace*{0mm} Text: \hspace*{2mm} {\it Romney will be a terrible president.}
\end{quote}
\vspace*{-1mm}
\noindent The tweet was posted during the 2012 US presidential campaign between Barack
Obama and Mitt Romney. Note that the text is negative in sentiment (and negative towards
Mitt Romney), but the tweeter is likely to be favorable towards the given target (Barack Obama). Also
note that one can be against Romney but not in favor of Obama, but in stance detection,
the goal is to determine which is more probable: that the author is in favour of, against,
or neutral towards the target.
\citeA{mohammad2017stance} analyzed manual annotations for stance and sentiment on the same data for a number of targets.

Even though about twenty teams participated in the first shared task on stance,
including some that used the latest recursive neural network models, none could do better
than a simple support vector machine baseline system put up by the organizers that used word and character n-gram features \cite{sobhani2016detecting}. Nonetheless, there has been considerable followup work since then \cite{yan2020efficient,xu2019adversarial,darwish2019unsupervised},
including new stance shared tasks for Catalan \cite{taule2017overview} and
work on jointly modeling stance and sentiment \cite{ebrahimi2016joint}.
See survey article by \citeA{kuccuk2020stance} for more details.

Automatically detecting stance has widespread applications in information retrieval, text
summarization, and textual entailment. One can also argue that stance detection is more useful
in commerce, brand management, public health, and policy making than simply identifying
whether the language used in a piece of text is positive or negative.



\subsection{Detecting Semantic Roles of Emotion}

The Theory of Frame Semantics argues that the meanings of most words can be understood in terms 
of a set of related entities and their relations \cite{fillmore1976frame,fillmore1982frame}.
For example, the concept of education usually involves a student, a teacher, a course, an institution,
duration of study, and so on.
The  set of related entities is called a {\it semantic frame} and 
the individual entities, defined in terms of the role they 
play with respect to the target concept, are called the {\it semantic roles}.
{\it FrameNet} \cite{Baker98} is a lexical database of English that records such semantic frames.\footnote{https://framenet.icsi.berkeley.edu/fndrupal/home}
Table 1 shows the FrameNet frame for emotions.  
Observe that the frame depicts various roles such as who is experiencing the emotion (the {\it experiencer}),
the person or event that evokes the emotion, and so on.
Information retrieval,
text summarization, and textual entailment
benefit from determining not just the emotional state but also from determining these semantic roles of emotion.

\begin{table}[t]
\centering
 \begin{tabular}{ll}
\hline \bf Role & \bf Description \\ \hline
 Core: &\\
$\;\;\;$ Experiencer  &the person that experiences or feels the emotion\\
$\;\;\;$ State        &the abstract noun that describes the experience\\
$\;\;\;$ Stimulus     &the person or event that evokes the emotional response\\
$\;\;\;$ Topic        &the general area in which the emotion occurs\\[3pt]
 Non-Core: &\\
$\;\;\;$ Circumstances    &the condition in which Stimulus evokes response\\
$\;\;\;$ Degree       &The extent to which the Experiencer's emotion deviates from the\\
 &norm for the emotion\\
$\;\;\;$ Empathy\_target &The Empathy\_target is the individual or individuals with which the\\
  &Experiencer identifies emotionally\\ 
$\;\;\;$ Manner    &Any description of the way in which the Experiencer experiences\\
                 &the Stimulus which is not covered by more specific frame elements\\
$\;\;\;$ Reason       &the explanation for why the Stimulus evokes an emotional response\\
\hline
\end{tabular}
  \vspace*{-1mm}
\caption{\label{tab:emoframe} The FrameNet frame for emotions.}
\end{table}

\citeA{Mohammad2015elec} created a corpus of tweets from
the run up to the 2012 US presidential elections, with annotations for valence, emotion,
stimulus, and experiencer. 
The tweets were also annotated for intent (to criticize, to support, to ridicule, etc.) and style (simple statement, sarcasm, hyperbole, etc.).
The dataset is made available for download.\footnote{Political Tweets Dataset: www.purl.org/net/PoliticalTweets}
They also show that emotion detection alone can fail to distinguish
 between several different types of intent. For example, the same emotion of disgust
 can be associated with the intents of `to criticize', `to vent', and `to ridicule'.
They also developed systems that automatically classify electoral tweets as per their emotion and purpose,
using various features that have traditionally been used in tweet classification, such as word and character ngrams,
word clusters,
valence association lexicons, and emotion association lexicons.
\citeA{Ghazi15} compiled FrameNet sentences that were tagged with the stimulus of certain emotions.
They also developed a statistical model to detect spans of text referring to the emotion stimulus. 

\section{Detecting Valence and Emotions in Sentences and Tweets}
\label{sec:autosystems}

Sentiment analysis systems have been applied to many different kinds of texts including customer reviews \cite{PangL08,Liu12,liu2015sentiment},
newspaper headlines \cite{Bellegarda10}, novels \cite{Boucouvalas02,JohnBX06,FranciscoG06,MohammadY11}, emails \cite{LiuLS03,MohammadY11},
blogs \cite{NeviarouskayaPI09,GenereuxE06,MihalceaL06}, and tweets \cite{Pak10,Agarwal11,Thelwall11,Mohammad12}. 
Often the analysis of documents and blog posts is broken down into determining the sentiment within each component sentence.
In this section, we discuss approaches for such sentence-level sentiment analysis.
Even though tweets may include more than one sentence, they are limited to 140 characters, and most are composed of just
one sentence. Thus we include here work on tweets as well.


Note also that the text genre often determines the kind of sentiment analysis suitable with it: for example,
customer reviews are more suited to determine how one feels about a product than to determine one's general emotional state;
personal diary type blog posts (and tweets) are useful for determining the writer's emotional state;\footnote{\citeA{Mohammad2015elec} showed that tweets predominantly convey the emotions of the tweeter.}
movie dialogues and fiction are well suited to identify the emotions of the characters (and not as much of the writer's emotional state); reactions to a piece of text (for example, the amount of likes, applause, or other reactions) are useful in studying
the reader's attitude towards the text; and so on.


\subsection{Machine Learning Algorithms}

Given some text and associated true emotion labels (commonly referred to as \textit{training data}), machine learning systems learn a model.
(The emotion labels are usually obtained through annotations performed by native speakers of the language.)
Then, given new previously unseen text, the model predicts its emotion label.
The performance of the model is determined through its accuracy on a held out \textit{test set} for which emotion labels are available as well.

For both the training and prediction phases, the sentences are first represented by a vector of numbers. These vectors can be a series of 0s and 1s or a series of real-valued numbers. Much of the work in the 1990s and 2000s represented sentences by carefully hand-engineered vectors, for example, whether the sentence has a particular word, 
 whether the word is listed as a positive term in the sentiment lexicon, 
 whether a positive word is preceded by a negation,
 the number of positive words in a sentence, and so on.
 The number of features can often be as large as hundreds of thousands.
 The features were mostly integer valued (often binary) and sparse, that is, a vector is composed mostly of zeroes and just a few non-zero integer values. 
 
However, the dominant methodology since the 2010s is to represent words and sentences using vectors made up of only a few hundred real-valued numbers.
These continuous word vectors, also called {\it embeddings},
are learned from deep neural networks (a type of machine learning algorithm). 
However, these approaches tend to require very large amounts of training data.
Thus, recent approaches to deep learning rely heavily on another learning paradigm called {\it transfer learning}.
The systems first learn word and sentence representations from massive amounts of raw text.
For example, BERT \cite{devlin-etal-2019-bert}---one of the most popular current approaches, is trained on the entire Wikipedia corpus (about 2,500 million words) and a corpus of books (about 800 million words).
Roughly speaking, the learning of the representations is driven by the idea that a good word or sentence representation is one that can be used to best predict the words or sentences surrounding it in text documents.
Once these sentence representations are learned, they may be tweaked by a second round of learning that uses a small amount of task-specific training data;
for example a few thousand emotion-labeled sentences. 

Influential work on low-dimensional continuous representations of \textit{words} includes models such as word2vec \cite{mikolov2013distributed}, GloVe \cite{pennington2014glove}, fastText \cite{bojanowski2016enriching}, and their variations \cite{collobert2011natural,le2014distributed,bojanowski2017enriching,mikolov-etal-2018-advances}.
\citeA{eisner-etal-2016-emoji2vec} presented work on representing emoji with embeddings using a large corpora of tweets. 
ELMo \cite{peters-etal-2018-deep} introduced a method to determine context-sensitive word representations, that is, instead of generating a fixed representation of words, it generates a different representation for every different context that the word is seen in.

Influential work on low-dimensional continuous representations of \textit{sentences} includes models such as ULM-FiT \cite{howard2018universal}, BERT \cite{devlin-etal-2019-bert}, XLNet \cite{yang2019xlnet}, GPT-2 \cite{radford2019language} and their variations such as DistilBERT \cite{sanh2019distilbert}, and RoBERTa \cite{liu2019roberta}.\footnote{Note that BERT and other related approaches can be used not only to generate sentence representations, but also to generate context-sensitive word representations (similar to ELMo).} 
Yet another variation, SentiBERT \cite{yin2020sentibert}, was developed to help in tasks such as determining the sentiment of larger text units from the sentiment of its constituents.  
A compilation of work on BERT and its variants is available online.\footnote{https://github.com/tomohideshibata/BERT-related-papers/blob/master/README.md\#multi-lingual} 


Despite the substantial dominance of neural and deep learning techniques in NLP research, more traditional machine learning frameworks,
such as linear regression, support vector machines, and decision trees, as well as more traditional mechanisms to represent text features,
such as word n-grams and sentiment lexicons, remain relevant. This is notably because of their simplicity, ease of use, interpretability,
and even their potential ability to complement neural representations and improve prediction accuracy.
Word and character ngrams are widely used as features in a number of text classification problems, and it is not surprising
to find that they are beneficial for sentiment classification as well.
Features from manually and automatically created sentiment lexicons such as word--valence association lexicons, 
word--arousal association lexicons, word--emotion association lexicons are also commonly used in conjunction with neural representations for
detecting emotions or for detecting affect-related classes such as personality traits and states of well-being \cite{MohammadB17wassa,SemEval2018Task1,chatterjee-etal-2019-semeval,agrawal-suri-2019-nelec}.
Examples of commonly used manually created sentiment lexicons are:
the General Inquirer (GI) \cite{Stone66}, 
the NRC Emotion Lexicon \cite{MohammadT10,MohammadY11},
the NRC Valence, Arousal, and Dominance Lexicon \cite{vad-acl2018},
and VADER \cite{hutto2014vader}.
Commonly used automatically generated sentiment lexicons include 
SentiWordNet (SWN) \cite{Esuli06}, Sentiment 140 lexicon \cite{MohammadKZ2013}, and NRC Hashtag Sentiment Lexicon \cite{MohammadK14}.
Other traditional text features include those derived from
parts of speech, punctuations (!, ???), word clusters, syntactic dependencies, negation terms ({\it no, not, never}), and word elongations ({\it hugggs, ahhhh}).
We will discuss manually and automatically generated emotion lexicons in more detail in Section \ref{sec:lex}.

\subsection{Detecting Valence}
\label{sec:valsystems}



There is tremendous interest in automatically determining valence in sentences and tweets through supervised machine learning systems. This is evident from the large number of research papers, textual datasets, shared task competitions, and machine learning systems developed for valence prediction over the last decade.\\ 

\noindent {\bf Shared Tasks}

\noindent A number of shared task competitions on valence
prediction 
have focused on tweets. These include the 2013, 2014, 2015, and 2016 SemEval shared tasks titled {\it Sentiment
Analysis in Twitter (SAT)} and the 2018 Shared task titled \textit{Affect in tweets}, 
the 2015 SemEval shared task {\it  Sentiment Analysis of Figurative
Language in Twitter}, and the SemEval-2020 shared task on Sentiment Analysis for Code-Mixed Social Media Text.
Shared tasks on other sources of text include
the SemEval-2019 Task 3: Contextual Emotion Detection in Text, 
the 2014, 2015, and 2016 SemEval shared tasks on {\it Aspect Based
Sentiment Analysis (ABSA)},  and the 2015 Kaggle competition {\it Sentiment Analysis on Movie
Reviews}.\footnote{http://aclweb.org/aclwiki/index.php?title=SemEval\_Portal\\
SemEval-2015 Task 10: http://alt.qcri.org/semeval2015/task10/\\
SemEval-2015 Task 11: http://alt.qcri.org/semeval2015/task11/\\
SemEval-2015 Task 12: http://alt.qcri.org/semeval2015/task12/\\ 
SemEval-2016 Task 4: http://alt.qcri.org/semeval2016/task4/\\
SemEval-2016 Task 5: http://alt.qcri.org/semeval2016/task5/\\
SemEval-2018 Task 1: https://competitions.codalab.org/competitions/17751\\
SemEval-2020 Task 3: https://competitions.codalab.org/competitions/19790\\
SemEval-2020 Task 9: https://competitions.codalab.org/competitions/20654\\
http://www.kaggle.com/c/sentiment-analysis-on-movie-reviews} 
Many of these tasks received submissions from more than 40 teams from universities, research labs, and companies across the world.
The SemEval-2018 Task 1: Affect in Tweets, is particularly notable, because it includes an array of subtasks on inferring both emotion classes and emotion intensity, provides labeled data for English, Arabic, and Spanish tweets, and for the first time in an NLP shared task, analyzed systems for bias towards race and gender \cite{SemEval2018Task1}. (We discuss issues pertaining to ethics and fairness further in the last section of this chapter.)\\

\noindent {\bf Systems}

\noindent The NRC-Canada system came first in
the 2013 and 2014 SAT competitions \cite{MohammadKZ13,Zhu_SemEval2014}, and the 2014 ABSA competition  \cite{Kiritchenko_SemEval2014}. 
The system is based on a supervised statistical text classification approach leveraging a variety of surface-form, semantic, and sentiment features.
Notably, it used word and character ngrams, manually created and automatically generated sentiment lexicons, parts of speech,
word clusters, and Twitter-specific encodings such as hashtags, creatively spelled words, and abbreviations ({\it yummeee, lol, etc}).
The sentiment features were primarily derived from novel high-coverage tweet-specific sentiment lexicons.
These lexicons were automatically generated from tweets with sentiment-word hashtags (such as {\it \#great, \#excellent}) and 
from tweets with emoticons (such as {\it :), :(}).
(More details about these lexicons in in Section \ref{sec:lex}).
\citeA{tang2014coooolll} created a sentiment analysis system 
that came first in the 2014 SAT
subtask on a tweets dataset.
It replicated many of the same features used in the NRC-Canada system, and additionally used features drawn from word embeddings.

First \citeA{Socher2013}, and then \citeA{le2014distributed}, obtained significant improvements in valence classification on a movie reviews
 dataset \cite{PangL08} using word embeddings.
 Work by \citeA{kalchbrenner2014convolutional,Zhu:2015:ICML,wang-etal-2016-dimensional}, 
 and others further explored the use
 recursive neural networks and word embeddings in sentiment analysis.
Recent work on Aspect Based Sentiment Analysis has also explored using recurrent neural network (RNNs) and long short-term memory (LSTM) (particularly types of neural network) \cite{tang-etal-2016-effective,chen-etal-2017-recurrent-attention,ruder-etal-2016-hierarchical}.
 
More recent works have explored transfer learning
for sentiment analysis \cite{peters-etal-2018-deep,radford2019language,devlin-etal-2019-bert}. 
Systems built for shared task competitions \cite{duppada-etal-2018-seernet,agrawal-suri-2019-nelec,huang-etal-2019-ana,abdou-etal-2018-affecthor} often make use of ensemble neural systems, most commonly building on pre-tarined BERT, ULMFit, GloVe, and word2vec, followed by fine-tuning the system
on the provided labeled training set.

Even though the vast majority of sentiment analysis work is on English datasets, there is growing research in
Chinese \cite{zhou2019sentiment,xiao2018using,wan2008using} 
and Arabic dialects \cite{dahou-etal-2016-word,el2017sentiment,gamal2019twitter}.
(Further details on Chinese sentiment analysis can be found in this survey article by \citeA{peng2017review};
Further details on Arabic sentiment analysis can be found in these survey articles by \citeA{al2019comprehensive,alowisheq2016arabic}.) 
There is some work on other European and Asian languages, but little or none on native African and indigenous languages from around the world.

\subsection{Automatically Detecting and Analyzing Emotions}
\label{sec:emosystems}



Labeled training data is a crucial resource required for building supervised machine learning systems.
Compiling datasets with tens of thousands of instances annotated for emotion is expensive in terms of time and money.
Asking annotators to label the data 
for a large
number of emotions, increases the cost of annotation.
Thus, focusing on a small number of emotions has the benefit of keeping costs down. 
Below we summarize work on compiling textual datasets labeled with emotions and  automatic methods for detecting emotions in text.
We group the work by the emotion categories addressed.
\begin{itemize}
\item {\bf Work on Ekman's Six and Plutchik's Eight}: 
Paul Ekman proposed a set of six basic emotions that include: joy, sadness, anger, fear, disgust, and surprise \cite{Ekman92,ekman2003unmasking}.
Robert Plutchik's set of basic emotions includes Ekman's six as well as trust and anticipation.
Figure \ref{fig:plutchik} shows how Plutchik arranges these emotions on a wheel such that
opposite emotions appear
diametrically opposite to each other.
Words closer to the center have higher intensity than those that are farther.
Plutchik also hypothesized how some secondary emotions can be seen as combinations of some of the basic (primary) emotions, for example,
optimism as the combination of joy and anticipation. See \citeA{plutchik1991emotions} for details about his taxonomy of emotions
created by primary, secondary, and tertiary emotions. 

Since the writing style and vocabulary in different sources, such as chat messages, blog posts, and newspaper articles,
can be very different, automatic systems that cater to specific domains are more accurate when trained on
data from the target domain.
\citeA{AlmRS05} annotated  22 Grimm fairy tales (1580 sentences) for Ekman emotions.\footnote{https://dl.dropboxusercontent.com/u/585585/RESOURCEWEBSITE1/index.html}
\citeA{SemEval2007} annotated newspaper headlines with intensity scores for each of the Ekman emotions,
referred to as the {\it Text Affect Dataset}.\footnote{http://web.eecs.umich.edu/$\sim$mihalcea/downloads.html\#affective}
\citeA{AmanS07} annotated blog posts with the Ekman emotions.
\citeA{Mohammad12port} experimented on the Text Affect and Aman datasets to show that word ngram features tend to be data and domain specific, and thus not useful when working on a test set from a different domain, whereas, emotion lexicon
features are more robust and apply more widely.

\citeA{Mohammad12} polled the Twitter API for tweets that have hashtag words such as {\it \#anger} and {\it \#sadness} 
corresponding to the eight Plutchik emotions.\footnote{http://saifmohammad.com/WebPages/lexicons.html}
He showed that these hashtag words act as good labels for the rest of the tweets,
and that this labeled dataset is just as good as the set explicitly annotated for emotions for emotion classification.
Such an approach to machine learning from pseudo-labeled data is referred to as {\it distant supervision}.
\citeA{suttles2013distant} used a similar distant supervision technique and collected tweets with emoticons, emoji, and hashtag words
corresponding to the Plutchik emotions. They developed an algorithm for binary classification of tweets along the four opposing Plutchik dimensions. 
\citeA{kunneman2014predictability} studied the extent to which hashtag words in tweets are predictive of the affectual state
of the rest of the tweet. They found that hashtags can vary significantly in this regard---some hashtags are strong indicators of the corresponding emotion
whereas others are not. Thus hashtag words must be chosen carefully when employing them for distant supervision.

As part of the SemEval-2018 Task 1: Affect in Tweets, \citeA{SemEval2018Task1}  
created labeled datasets of English, Arabic, and Spanish tweets. The labels include: emotion intensity scores,
emotion intensity ordinal classification,
and emotion classes, in addition to valence annotations. 
As part of the SemEval-2019 Task 3: EmoContext Contextual Emotion Detection in Text, 
\citeA{chatterjee-etal-2019-semeval} provide a dataset of dialogues between a user and computer agents that are annotated for the emotions of the user.


 \begin{figure}[t]
 \begin{center}
 \includegraphics[width=0.48\columnwidth]{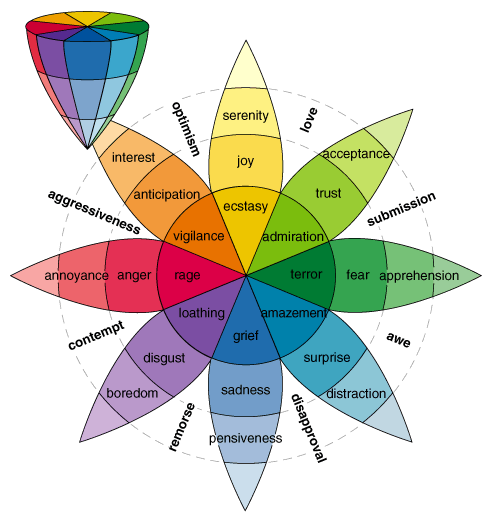}
 \end{center}
 \caption{Set of eight basic emotions proposed in \citeA{Plutchik80}.}
 \label{fig:plutchik}
 \end{figure}


\item {\bf Work on Other Small Sets of Emotions}:
The ISEAR Project asked 3000 student respondents to report situations in which they had experienced
joy, fear, anger, sadness, disgust, shame, or
guilt.\footnote{http://emotion-research.net/toolbox/\\toolboxdatabase.2006-10-13.2581092615}
\citeA{thomas2014synthesized} applied supervised machine learning techniques on the ISEAR dataset for
7-way emotion classification.  \citeA{NeviarouskayaPI09} collected 1000
sentences from the Experience Project webpage and  manually annotated them for fourteen affectual
categories.\footnote{www.experienceproject.com} Experience Project is a portal where users share
their life experiences. These shared texts are usually rife with emotion.


\citeA{Bollen2009} analyzed 9,664,952 tweets posted in the second half of 2008 
using Profile of Mood States (POMS) \cite{mcnair1989profile}.
POMS is a psychometric instrument that measures the mood states of tension, depression, anger, vigor, fatigue, and confusion.



\item {\bf Distant Supervision Datasets:} Distant supervision techniques proposed in \citeA{Mohammad12,Purver12,Wang12,suttles2013distant}, and others
have opened up the critical bottleneck of creating instances labeled with emotion categories. Thus,
now, labeled data can be created for any emotion for which there are sufficient number of tweets
that have the emotion word as a hashtag. \citeA{MohammadK14} collected tweets with hashtags corresponding to
around 500 emotion words as well as positive and negative valence. They used these tweets to identify words associated
with each of the 500 emotion categories, which were in turn used as features in a task of
automatically determining personality traits from stream-of-consciousness essays.
They show that using features from 500 emotion categories significantly improved performance
over using features from just the Ekman emotions.
\citeA{pool-nissim-2016-distant} collected public Facebook posts and their associated user-marked
emotion reaction data from news and entertainment companies to develop an emotion classification system. 
\citeA{felbo-etal-2017-using} collected millions of tweets with emojis and developed neural transfer learning
models from them that were applied to sentiment and emotion detection tasks.
\citeA{abdul-mageed-ungar-2017-emonet} collected about 250 million tweets pertaining to 24 emotions and 665 hashtags
and developed a Gated Recurrent Neural Network (GRNN) model for emotion classification.

\item {\bf Work on Emotion-Labeled Datasets in Languages Other than English:}
\citeA{wang2014segment} annotated Chinese news and blog posts with the Ekman emotions. 
Wang also translated Alm's fairy tales dataset into Chinese.
\citeA{quan2009construction} created a blog emotion corpus in Chinese called the
{\it Ren-CECps Corpus}. The sentences in this corpus are annotated with eight emotions:
 expectation, joy, love, surprise, anxiety, sorrow, anger, and hate. Sentences not associated
 with any of these eight categories are marked as neutral.
The corpus has 1,487 documents, 11,255 paragraphs, 35,096 sentences, and 878,164 Chinese words.

The 2013 Chinese Microblog Sentiment Analysis Evaluation (CMSAE) compiled a dataset of
posts from Sina Weibo (a  popular Chinese
microblogging service) annotated with 
seven emotions: anger, disgust, fear, happiness, like, sadness and surprise.\footnote{http://tcci.ccf.org.cn/conference/2013/pages/page04\_eva.html}
If a post has no emotion, then it is labeled as {\it none}. The training set contains 4000 instances (13252 sentences).  
The test dataset
contains 10000 instances (32185 sentences). 
Recent works on Chinese emotion classification using transfer learning and neural networks include 
\citeA{yu2018improving,matsumoto2018transfer}.

Notable Arabic emotion datasets include the tweet datasets from
the shared task Semeval-2018 task 1: Affect in Tweets \citeA{SemEval2018Task1} 
and other works such as \citeA{badarneh2018fine,al2017emotional}.
\citeA{hussien2016emoticons} analyzed the effectiveness of using emoticons for distant supervision in Arabic tweets.


\item {\bf Multimodal Datasets:} There is new work on developing word representations not only from text corpora
but also from collections of images that the words are associated with \cite{kiela2014learning,hill2014learning,lazaridou2014wampimuk}. (An image and a word
can be considered associated if the caption for the image has the word, the image is a
representation of the concept the word refers to, etc.)
This bridges the gap between text and vision, allowing the exploration of new applications such as
automatic image captioning \cite{karpathy2014deep,kiros2014unifying}.
\citeA{mohammad-kiritchenko-2018-wikiart} created a dataset of more than 4,000 pieces of art (mostly paintings) that has annotations for emotions evoked in the observer by the painting (image and title), the title alone, and the image alone. The pieces of art are selected from WikiArt.org's collection for four western styles (Renaissance Art, Post-Renaissance Art, Modern Art, and Contemporary Art). In addition, there is a large body of work on analyzing emotions
through videos that analyzes facial expressions, gestures, text transcripts, and vocal features.  
(See survey by \citeA{marechal2019survey}.) 
Such multi-modal representations of emotions are useful not only to determine the emotional states of entities more accurately but also in tasks such as 
captioning images or audio for emotions and even 
generating text that is affectually suitable for a given image or audio sequence.

\end{itemize}


\noindent As seen above, there are a number of datasets where sentences are manually labeled for emotions.
They have helped improve our understanding of how people use language to convey emotions 
and they have helped in developing supervised machine learning emotion classification systems.
 Yet, a number of questions remain unexplored. For example,
 to what extent do people convey emotions without using explicit emotion words?
 are some emotions more basic than others? is there a taxonomy of emotions?
 are some emotions indeed combinations of other emotions (optimism as the combination of joy and anticipation)?
 can data labeled for certain emotions be useful in detecting certain other emotions?
 and so on.

\subsection{Detecting Subjectivity}
One of the earliest problems tackled in sentiment analysis is that of detecting subjective language \cite{wiebe2004learning,wiebe2005creating}.
For example, sentences can be classified as subjective (having opinions and attitude) or objective (containing facts).
This has applications in question answering, information retrieval, paraphrasing,
and other natural language  applications where it is useful to separate factual statements from speculative or affectual ones.
For example, if the target query is ``{\it what did the users think of iPhone 5's screen?}", then the question answering system (or information retrieval system)
should be able to distinguish between sentences such as {\it ``the iPhone has a beautiful touch screen"} and
sentences such as {\it ``iPhone 5 has 326 pixels  per inch"}.
Sentences like the former which express opinion about the screen should be extracted.
On the other hand, if the user query is ``{\it what is iPhone 5's screen resolution?}", then sentences such as the latter (referring to 326 pixels per inch) are more relevant.
(See \citeA{wiebe2011finding} for work on using subjectivity detection in tandem with techniques for information extraction.)
It should be noted, however, that if a sentence is objective, then it does not imply that the sentence is necessarily true.
It only implies that the sentence does not exhibit the speaker's private state (attitude, evaluations, and emotions).
Similarly, if a sentence is subjective, that does not imply that it lacks truth.

A number of techniques have been proposed to detect subjectivity using patterns of word usage, identifying certain kinds of adjectives, 
detecting emotional terms, and occurrences of certain discourse connectives
\cite{hatzivassiloglou2000effects,riloff-wiebe-2003-learning,wiebe2004learning}. 
Opinion Finder is a popular freely available subjectivity systems \cite{wilson2005opinionfinder}.\footnote{http://mpqa.cs.pitt.edu/opinionfinder/}
However, there has been very little work on subjectivity detection in the later half of 2010s;
likely because of the greater interest in modeling sentiment and emotion directly as opposed to through the framework of subjectivity.





\section{Capturing Term--Sentiment Associations}
\label{sec:lex}

The same word can convey different sentiment in different contexts. For example,
the word {\it unpredictable} is negative in the context of automobile steering,
but positive in the context of a movie script.
Nonetheless, many words have a tendency to convey the same sentiment in a large majority
of the contexts they occur in. For example, {\it excellent} and {\it cake} are positive in most
usages whereas {\it death} and {\it depression} are negative in most usages.
These majority associations are referred to as {\it prior associations}.
Sentiment analysis systems benefit from knowing these prior associations of words and phrases.
Thus, lists of term--sentiment associations have been created by manual annotation.
These resources tend to be small in coverage because manual annotation is expensive
and the number of words and phrases for a language can run into hundreds of thousands.
This has led to the development of automatic methods that extract large lists of term--sentiment 
associations from text corpora using manually created lists
as seeds.
We describe work on manually creating and automatically generating term--sentiment associations in the 
sub-sections below. (See \citeA{mohammad2020practical} for a discussion of the ethical considerations in the use of emotion and sentiment lexicons.)


\subsection{Manually generated term-sentiment association lexicons}
\label{sec:manlex}



Even though some small but influential lexicons such as the General Inquirer \cite{Stone66} and ANEW \cite{bradley1999affective}
were created decades earlier, the 2010s saw notable progress in two salient areas: crowdsourced emotion lexicons
and reliable real-valued annotations. We discuss the two in the sub-sections below, followed by a list of prominent emotion lexicons and their details.

\subsubsection{Crowdsourced Emotion Lexicons}

Crowdsourcing involves breaking down a large task into small independently solvable units, distributing the units through
the Internet or some other means, and getting a large number of people to solve or annotate the units.
The requester specifies the compensation that will be paid for solving each unit.
In this scenario, the annotators are usually not known to  the requester and usually do not all have the same
academic qualifications.
Natural language tasks are particularly well-suited for crowdsourcing because even though computers
find it difficult to understand language, native speakers of a language do not usually need extensive training
to provide useful annotations such as whether a word is associated with positive sentiment.
Amazon Mechanical Turk and CrowdFlower are two commonly used crowdsourcing platforms.\footnote{https://www.mturk.com/mturk/welcome\\http://www.crowdflower.com}
They allow for large scale annotations, quickly and inexpensively.
However, one must define the task carefully to obtain annotations of
high quality. Checks must be placed to ensure
that random and erroneous annotations are discouraged, rejected, and re-annotated.

The NRC Emotion Lexicon (EmoLex) was the first emotion lexicon created by crowdsourcing. It includes
entries for $\sim$14k English terms. Each entry includes ten binary scores (0 or 1) indicating no association or association with eight basic emotions as well as positive and negative sentiment \cite{MohammadT13,MohammadT10}.\footnote{http://saifmohammad.com/WebPages/NRC-Emotion-Lexicon.htm} 
For each word, annotators were first asked a simple word choice question where one of the options (the correct answer) is taken from a thesaurus, and the remaining options are made up of randomly selected words. 
This question acts as a mechanism to check whether the annotator knew the meaning of the word and was answering questions diligently.
The question also serves a mechanism to bias the annotator response to a particular sense of the word.
(Different senses of a word can convey differing emotions.)
The resulting lexicon has entries for about 25K word senses.
A word-level lexicon (for $\sim$14K words) was created by taking the union of the emotions associated with each of the 
senses of a word.

Other examples of crowdsourced emotion lexicons include the NRC Valence, Arousal, and Dominance Lexicon \cite{vad-acl2018}
and the NRC Affect Intensity Lexicon \cite{LREC18-AIL}. However, these required real-valued scores (and not just binary
scores), which entail additional challenges. We discuss them next.




\subsubsection{Real-valued Sentiment Scores from manual Annotations}
\label{sec:realval}
Words have varying degrees of associations with sentiment categories. 
This is true not just for comparative and superlative adjectives and adverbs (for example, \textit{worst} 
is more negative than \textit{bad}) 
but also for other syntactic categories. For example,
most people will agree that {\it succeed} is more positive (or less negative) than {\it improve}, and
{\it fail} is more negative (or less positive) than {\it deteriorate}.
Downstream applications benefit from knowing not only whether a word or phrase in positive or negative (or associated with some emotion category),
but also from knowing the strength of association.
However, for people, assigning a score indicating the degree of sentiment is not natural.
Different people may assign different scores to the same target item, and it is hard for even the same annotator
to remain consistent when annotating a large number of items.
In contrast, it is easier for annotators to determine whether one word is more positive (or more negative) than the
other.
However, the latter requires a much larger number of annotations than the former
(in the order of $N^2$, where $N$ is the number of items to be annotated).

{\it Best--worst scaling (BWS)} is an annotation scheme that retains the comparative aspect of annotation while
still requiring only a small number of annotations  \cite{Louviere_1991,louviere2015best}.
It has its basis in the mathematical choice modeling and psycho-physics.
Essentially, the annotator is presented with four (or five) items and
asked which is the most (say, most positive) and which is the least (say, least positive). By answering just
these two questions five out of the six inequalities are known.
If the respondent says that $A$ is most
positive and $D$ is least positive, then we know:
\begin{quote}
     $A > B, A > C, A > D, B > D, C > D$
\end{quote}
Each of these BWS questions can be presented to multiple annotators.  

The responses to
the BWS questions can then be easily translated into a ranking of all the terms and
also a real-valued score for all the terms \cite{Orme_2009}.  
Kiritchenko and Mohammad \shortcite{KiritchenkoM2017bwsvsrs} showed through empirical experiments that BWS produces more reliable and more discriminating scores than those obtained using rating scales.\footnote{See \citeA{maxdiff-naacl2016,KiritchenkoM2017bwsvsrs} for further details on BWS and its use in NLP applications.}

BWS was used for obtaining annotations of relation similarity of pairs of items by \cite{jurgens-EtAl:2012:STARSEM-SEMEVAL}
in a SemEval-2012 shared task.
\citeA{Kiritchenko2014} used BWS to create a dataset of 1500 Twitter terms
with real-valued sentiment association scores.

Real-valued valence scores obtained through BWS annotations were used in subtask E of the 2015
SemEval Task {\it Sentiment Analysis in Twitter} \cite{rosenthal2015semeval} to evaluate automatically
generated Twitter-specific valence lexicons.
Datasets created with the same approach were used in a 2016 Task {\it Determining sentiment intensity of English and Arabic phrases}
to evaluate both English and Arabic automatically generated sentiment lexicons.

BWS was also used to create the NRC Valence, Arousal, and Dominance Lexicon (NRC-VAD) \cite{vad-acl2018}
and the  NRC Emotion Intensity Lexicon (NRC-EIL) \cite{LREC18-AIL}.
With over 20K entries, NRC-VAD is the largest manually created emotion lexicon. 
\citeA{vad-acl2018} also analyzed the annotations to show how different demographic groups,
such as men and women, people with different personality traits, etc., perceive
valence, arousal, and dominance differently.
NRC-EIL provides intensity scores for the words and emotions in the NRC Emotion Lexicon.
\citeA{mohammad2018understanding} analyzed the entries across NRC-VAD and NRC-EIL
to show that while the positive  and negative emotions are clearly separated 
across the valence dimension, emotions such as anger, fear, and disgust do not occur
in clearly separate regions of the VAD space.

\subsubsection{Large, Influential, Manually Created Lexicons}
The earliest emotion lexicons such as the General Inquirer and ANEW capture words that denotate (express) an emotion. These lexicons tend to be small. Starting with the NRC Emotion lexicon, we see work on capturing not just words that denotate an emotion, but also those that connotate an emotion. For example, the word {\it party} does not denotate joy, but is associated with, or connotates, joy. 
Below are some of the most widely used sentiment and emotion lexicons. They are ordered by size. 

\begin{itemize}

\item {\bf NRC Valence, Arousal, and Dominance Lexicon (NRC-VAD):} \textbf{$\sim$20K English terms} annotated for real-valued scores of \textbf{valence, arousal and dominance} \cite{vad-acl2018}.\footnote{http://saifmohammad.com/WebPages/nrc-vad.html} 
Automatic translations of the words in over 100 languages included.

\item {\bf NRC Emotion Lexicon (EmoLex)}: \textbf{$\sim$14k English terms} with binary annotations (0 or 1) indicating no association or association with \textbf{eight basic emotions} (those proposed by \citeA{Plutchik80}) as well as for positive and negative sentiment.\footnote{http://saifmohammad.com/WebPages/NRC-Emotion-Lexicon.htm} 
Automatic translations of the words in overl 100 languages included \cite{MohammadT10,MohammadT13}.

\item {\bf Warriner et al.\@ Lexicon:} \textbf{$\sim$14K English terms} annotated for \textbf{valence, arousal, and dominance} (real-valued scores) \cite{warriner2013norms}.\footnote{http://crr.ugent.be/archives/1003}

\item {\bf NRC Emotion Intensity Lexicon:} \textbf{$\sim$10K English terms} annotated for intensity (real-valued) scores corresponding to \textbf{eight basic emotions} (anger, anticipation, disgust, fear, joy, sadness, surprise, and trust) \cite{LREC18-AIL}.\footnote{http://saifmohammad.com/WebPages/AffectIntensity.htm}

\item {\bf MPQA Subjectivity Lexicon:} \textbf{$\sim$8,000 English terms} annotated for \textbf{valence} (strongly positive, weakly positive, strongly negative and weakly negative)  \cite{Wilson05}.\footnote{https://mpqa.cs.pitt.edu/lexicons/subj\_lexicon/}
Includes the words and annotations from the General Inquirer and
other sources. 

\item {\bf Hu and Liu Lexicon:} \textbf{$\sim$6,800 English terms} from customer reviews that are annotated for \textbf{valence} (positive, negative) \cite{Hu04}.\footnote{http://www.cs.uic.edu/$\sim$liub/FBS/opinion-lexicon-English.rar} 

\item {\bf General Inquirer (GI):} \textbf{$\sim$3,600 English terms} annotated for associations with various semantic categories including \textbf{valence} \cite{Stone66}.\footnote{http://www.wjh.harvard.edu/~inquirer/\\Note: \citeA{Stone66} use the term evaluativeness instead of valence.}
These include about 1500 words from the Osgood study.

\item {\bf AFINN:}  \textbf{$\sim$2,500 English terms} annotated for \textbf{valence} \cite{AFINN}. The ratings range from -5 (most negative) to +5 (most positive)
in steps of 1.\footnote{http://www2.imm.dtu.dk/pubdb/views/publication\_details.php?id=6010\\
Note: The publication describing AFINN does not state whether the name is an abbreviation.}

\item {\bf Linguistic Inquiry and Word Count (LIWC) Dictionary:} \textbf{$\sim$1,400 English terms} manually identified to denotate the \textbf{affect categories}: positive emotion, negative emotion, anxiety, anger, and sadness \cite{pennebaker2015development}.\footnote{LIWC also includes other words associated with categories such as personal pronouns and biological concern. The LIWC dictionary also comes with software to analyze text and predict the psychological state of the writer by identifying repeated use of words from various categories.}

\item {\bf The Affective Norms for English Words (ANEW):} \textbf{$\sim$1,000 English terms} annotated for \textbf{valence, arousal, and dominance}  \cite{bradley1999affective}.\footnote{http://csea.phhp.ufl.edu/media/anewmessage.html}

\end{itemize}

\noindent Emotion lexicons such as those listed above provide simple and effective means to analyze and draw inferences from large amounts of text. They are also used by machine learning systems to improve prediction accuracy, especially when the amount of training data is limited. Large emotion lexicons are also widely used in digital humanities, literary analyses, psychology, and art for a number purposes.


\subsection{Automatically inducing term--sentiment association lexicons}
\label{sec:autolex}
Automatic methods for capturing word--sentiment associations can quickly learn
associations for hundreds of thousands of words, and even for sequences of words.
They can learn associations that are relevant to a particular domain \cite{chetviorkin2014two,hamilton2016inducing}.
For example, when the algorithm is applied on a text of movie reviews,
the system can learn that {\it unpredictable} is a positive term in this domain (as in {\it unpredictable story line}), but when
applied to auto reviews, the system can learn that  {\it unpredictable} is a negative term (as in {\it unpredictable steering}).
They also tend to have higher coverage (include more terms) than manually created lexicons. Notably, the coverage of domain specific terms is often much better in automatically generated lexicons than general purpose manually created lexicons. For example, a sentiment lexicon that is generated from a large number of freely available tweets, such as the Sentiment140 lexicon \citeA{MohammadSemEval2013}, has a number of entries for Twitter-specific languages such as emoticons ({\it :-)}), hashtags ({\it \#love}), conjoined words ({\it loveumom}), and creatively spelled words ({\it yummeee}).
Similarly, sentiment lexicons, such as SenTrop \cite{hamilton2016inducing}, can be generated that are appropriate for different historical time periods.\footnote{Over time, words can change meaning and also sentiment. For example, words such as terrific and lean originally had negative connotations, but now they are largely considered to have positive connotations.} 
Automatic methods can also be used to create separate lexicons for words found in negated context and those found in
affirmative context \cite{Kiritchenko2014}; the idea being that the same word contributes to sentiment
differently depending on whether it is negated or not.  These lexicons can contain sentiment
associations for hundreds of thousands of unigrams, and even larger textual units such as bigrams and trigrams.  

Methods for automatically inducing sentiment lexicons rely on three basic components:
a large lexical resource,
a small set of of seed positive and negative terms (sometimes referred to as paradigm words),
and a method to propagate or induce sentiment scores for the terms in the lexical resource.
Commonly used lexical resources are: semantic networks such as WordNet   
\cite{Hatzivassiloglou97,baccianella2010sentiwordnet}, thesauri \cite{Mohammad09}, 
and Wikipedia \cite{chen-skiena-2014-building};
text corpora such as the world wide web data \cite{TurneyL03} and collections of tweets \citeA{MohammadK13,abdul-mageed-ungar-2017-emonet}.
A popular variant in recent years is to make use of word embeddings (which themselves are generated from extremely large text corpora) as the primary lexical resource
 \cite{faruqui-etal-2015-retrofitting,yu-etal-2017-refining}.

Seed words are compiled manually or through existing
manually created sentiment lexicons.
\citeA{TurneyL03} and \citeA{hamilton2016inducing} identified small lists of positive and negative words that are largely
monosemous and have stable sentiments in different contexts and time periods.
\citeA{Go09} used emoticons as seed terms to induce valence of words in tweets.
\citeA{MohammadSemEval2013} used hashtags ({\it \#angry, \#sad, \#good, \#terrible,} etc.) and emoticons (:), :() to induce valence and emotion association scores for eight basic emotions and positive negative sentiment.
\citeA{MohammadK13,abdul-mageed-ungar-2017-emonet} used hashtags to induce emotion associations for dozens of different emotions.

The central idea in methods to induce sentiment lexicons is to propagate the sentiment information from the 
seed terms to other terms in the lexical resource. This method of propagation can make use of simple word--seed
co-occurrence in text \cite{TurneyL03,MohammadSemEval2013} or
any of the graph-propagation algorithms in a semantic network \cite{baccianella2010sentiwordnet}.

Word embeddings, which are induced from text corpora, are designed to produce word representations such that if two words are close in meaning, then their representations are close to each other (in vector space).
They rely on the idea that words that are close in meaning also occur in similar contexts---distributional hypothesis \cite{Firth57,Harris68}.
Methods to induce sentiment lexicons from word embeddings propagate sentiment labels or scores from the seed
words to other words in the vector space in such a way that the sentiment score of a word is influenced strongly by the sentiment scores of the words closest to it.
However, antonymous words are known to co-occur with each other more often than random chance \cite{Cruse86,Fellbaum95,MohammadDH08}.
Thus, to some extent, antonymous words tend to occur close to each other in word embeddings, making them somewhat problematic in the propagation of sentiment.
Thus, several approaches have been proposed that make use of word embeddings (as the primary lexical resource) and some sentiment signal (star rating of reviews or seed set of sentiment words). These approaches are of two kinds: (1) those that learn word embeddings with an additional objective function to keep antonymous and differing sentiment word pairs farther away from each other  \cite{maas-etal-2011-learning,tang2014building,Ren2016ImprovingTS}; and (2) those that modify (refine/retrofit) word embeddings post-hoc \cite{labutov-lipson-2013-embedding,faruqui-etal-2015-retrofitting,hamilton2016inducing,yu-etal-2017-refining}. 

\section{Modeling the impact of sentiment modifiers}
\label{sec:mods}

Negation words (e.g., {\it  no, not, never, hardly, can't, don't}), modal verbs (e.g., {\it can, may, should, must}), degree adverbs (e.g., {\it almost, nearly, seldom, very}) and other modifiers impact the sentiment of the term or phrase they modify.
\citeA{SCL-NMA} manually annotated a set of phrases that include negators, modals, degree adverbs, and their combinations. Both the phrases and their constituent content words are annotated with real-valued scores of sentiment intensity using the technique Best–Worst Scaling. 
They measured the effect of individual modifiers as well as the average effect of the groups of modifiers on overall sentiment.
They found that the effect of modifiers varies substantially among the members of the same group. Furthermore, each individual modifier can affect the modified sentiment words in different ways.
Their lexicon, \textit{Sentiment Composition Lexicon of Negators, Modals, and Adverbs (SCL-NMA)}, was used as an official test set in the SemEval-2016 shared Task \#7: Detecting Sentiment Intensity of English and Arabic Phrases. The objective of that task was to automatically predict sentiment intensity scores for multi-word phrases. Below we outline some focused work on individual groups of sentiment modifiers.

\subsection{Negation}
\citeA{morante2012modality} define negation to be ``a grammatical category that allows the
changing of the truth value of a proposition''. Negation is often expressed through the
use of negative signals or negator words such as {\it not} and {\it never}, and it can significantly
affect the sentiment of its scope. Understanding the impact of negation on sentiment improves
 automatic detection of sentiment.

Automatic negation handling involves identifying a negation word such as {\it not}, determining the scope of negation (which words are affected
 by the negation word), and finally appropriately capturing the impact of the negation.
 (See work by \citeA{Jia09,Wiegand10,LapponiRO12} for detailed analyses of negation handling.)
 Traditionally, the negation word is determined from a small hand-crafted list \cite{Taboada_2011}.
 The scope of negation is often assumed to begin from the word following the negation word until the next
 punctuation mark or the end of the sentence \cite{Polanyi04,Kennedy05}.
 More sophisticated methods to detect the scope of negation through semantic parsing have also been proposed \cite{Li10}.

 Earlier works on negation handling employed simple heuristics such as flipping the polarity
 of the words in a negator's scope \cite{Kennedy05} or changing the degree of sentiment of the
 modified word by a fixed constant \cite{Taboada_2011}. 
\citeA{Zhu2014} show that these simple heuristics fail to capture the true impact of negators
on the words in their scope. They show that negators tend to often make positive words negative
(albeit with lower intensity) and make negative words less negative (not positive).
Zhu et al. also propose certain embeddings-based recursive neural network models
to capture the impact of negators more precisely.
 As mentioned earlier, \citeA{Kiritchenko2014} capture the impact of negation by creating separate
 sentiment lexicons
 for words seen in affirmative context and those seen in negated contexts.
 These lexicons are generated using co-occurrence statistics of terms in affirmative context with sentiment signifiers
 such as emoticons and seed hashtags (such as {\it \#great, \#horrible}), and separately for 
 terms in negated contexts with sentiment signifiers.
  They use a hand-chosen list of negators and determine scope to be starting from the negator
  and ending at the first punctuation (or end of sentence).


\subsection{Degree Adverbs, Intensifiers, Modals}
Degree adverbs such as {\it barely, moderately}, and {\it slightly} quantify the extent or amount of the predicate.
Intensifiers such as {\it too} and {\it very} are modifiers that do not change the propositional content (or truth value) of the predicate they modify,
but they add to the emotionality.
However, even linguists are hard-pressed to give out comprehensive lists of degree adverbs and intensifiers.
Additionally, the boundaries between degree adverbs and intensifiers can sometimes be blurred,
and so it is not surprising that the terms are occasionally used interchangeably.
Impacting propositional content or not, both degree adverbs and intensifiers impact the sentiment of the predicate,
and there is some work in exploring this interaction
\cite{zhang2008polarity,wang2012bootstrapping,xu2008learning,lu2010cityu,taboada2008extracting}.
Most of this work focuses on identifying sentiment words by bootstrapping over patterns involving degree adverbs and intensifiers.
Thus several areas remain unexplored,
such as
identifying patterns and regularities in how different kinds of degree adverbs and intensifiers impact sentiment,
ranking degree adverbs and intensifiers in terms of how they impact sentiment, and
determining when (in what contexts) the same modifier will impact sentiment differently than its usual behavior.

Modals are a kind of auxiliary verb used to convey the degree of confidence, permission, or obligation.
Examples include {\it can, could, may, might, must, will, would, shall,} and {\it should}.
The sentiment of the combination
of the modal and an expression can be different from the sentiment of the expression alone. For example,
{\it cannot work} is less positive than {\it work} or {\it will work} ({\it cannot} and {\it will} are modals).
Thus handling modality appropriately can greatly improve automatic sentiment analysis systems.

%
%
%
%

\section{Sentiment in Sarcasm, Metaphor, and other Figurative Language}
\label{sec:figurative}

There is growing interest in detecting figurative language, especially irony and
 sarcasm \cite{carvalho2009clues,reyes2013multidimensional,veale2010detecting,filatova2012irony,gonzalez2011identifying}.
In 2015, a SemEval shared task was organized on detecting sentiment in tweets rich in metaphor and irony (Task 11).\footnote{The proceedings
will be released later in 2015.} Participants were asked to determine the degree of sentiment for each tweet where the score is a real number in the range
from -5 (most negative) to +5 (most positive). One of the characteristics of the data is that most of the tweets are negative; thereby
suggesting that ironic tweets are largely negative.
The SemEval 2014 shared task Sentiment Analysis in Twitter \cite{SemEval2014task9} had a separate test set involving sarcastic tweets.
Participants were asked {\it not} to train their system on sarcastic tweets, but rather apply their regular sentiment system on this new test set;
the goal was to determine performance of
regular sentiment systems on sarcastic tweets. It was observed that the performances dropped by about 25 to 70 percent,
thereby showing that systems must be adjusted if they are to be applied to sarcastic tweets.

It is generally believed that a metaphor tends to have an emotional impact, and thus it is not surprising that they are used widely in language. 
However, their inherent non-literality can pose a challenge to sentiment analysis systems. Further, the mechanisms through which metaphors convey emotions are not well understood. 
\citeA{MohammadST16} presented a study comparing the emotionality of metaphorical English expressions with that of their literal counterparts. They found that metaphorical usages are, on average, significantly more emotional than literal usages. They also showed that this emotional content is not simply transferred from the source domain into the target, but rather is a result of meaning composition and interaction of the two domains in the metaphor.
 \citeA{ho2016metaphors} analyses a text of English financial analysis reports for the use of emotional metaphors.
\citeA{rai2019understanding} show how models for understanding metaphors benefit from sensing the associated emotions.
See work by \cite{dankers2019modelling} on joint models for detecting metaphor and sentiment.
\cite{su2019chinese} developed a method to detect sentiment of Chinese metaphors.

\citeA{Balahur2010SentimentAI} and \citeA{liu-etal-2017-idiom} showed that automatic sentiment analysis systems often have difficulties in dealing with idioms.
\citeA{Williams2015TheRO} and \citeA{spasic2017idiom} show that sentiment analysis in English texts can be improved using idiom-related features.
\citeA{Williams2015TheRO} and \citeA{jochim-etal-2018-slide} annotated lists of 580 and 5000 English idioms, respectively, that are manually annotated for sentiment.
\citeA{passaro2019idioms} collected valence and arousal ratings for 45 Italian verb-noun idioms and 45 Italian non-idiomatic verb-noun pairs. 

We found little to no work exploring automatic sentiment detection in
hyperbole, understatement, rhetorical questions, and other creative uses of language.

\section{Multilingual and Crosslingual Sentiment Analysis}

A large proportion of research in sentiment analysis, and natural language processing in general, has focused on English. Thus for languages other than English there are fewer and smaller
resources (sentiment lexicons, emotion-annotated corpora, etc.).
This means that automatic sentiment analysis systems in other languages tend to be less accurate than
their English counterpart.
Work in \textit{Multilingual Sentiment Analysis} aims at building multilingual affect-related resources, as well as developing (somewhat) language independent approaches for sentiment analysis---approaches that can be applied to wide variety of resource-poor languages. A common approach to sentiment analysis in a resource-poor \textit{target language} is to leverage powerful resources from another \textit{source language} (usually English) via translation. This type of work is usually referred to as {\it Crosslingual Sentiment Analysis}. 

Work in multilingual and crosslingual sentiment analysis includes that on 
leveraging source language sentiment annotated corpora \cite{Pan2011CrossLingualSC,arabicSA,chen-etal-2018-adversarial}; 
unlabeled bilingual parallel data \cite{meng-etal-2012-cross},
source language sentiment lexicons \cite{Mihalcea07learningmultilingual,arabicSA2}, 
multilingual WordNet \cite{bobicev2010emotions,cruz2014building}, 
multilingual word and sentiment embeddings (learned from sentiment annotated data in source and target language) \cite{zhou-etal-2016-cross,feng2019learning}, 
etc. 



\citeA{duh-etal-2011-machine} presented an opinion piece on the challenges and opportunities of using 
automatic machine translation for crosslingual sentiment analysis, and notably suggest the need for new domain adaptation techniques.
\citeA{balahur2014comparative} conducted a study to assess the performance of 
sentiment analysis techniques on machine-translated texts.  
\citeA{arabicSA} and \citeA{arabicSA2}
conducted experiments to determine loss in sentiment predictability when they
translate Arabic social media posts into English, manually and automatically. As a benchmark, they use manually 
determined sentiment of the Arabic text. They show that
an English sentiment analysis system has only a slightly lower accuracy on the English translation of Arabic text 
as compared to the accuracy of an Arabic sentiment analysis system on the original (untranslated) Arabic text.
They also showed that automatic Arabic translations of English valence lexicons improve accuracies
of an Arabic sentiment analysis system.\footnote{The translated lexica and corpora for Arabic sentiment analysis: www.purl.org/net/ArabicSA}

Some of the areas less explored in the realm of multilingual sentiment analysis include:
how to translate text so as to preserve the degree of sentiment in the source text;
how sentiment modifiers
such as negators and modals differ in function across languages;
understanding how automatic translations differ from manual translations in terms of sentiment;
and how to translate figurative language without losing its affectual gist.


\section{Applications}

Automatic detection and analysis of affectual categories in text has wide-ranging applications. Below we list some key directions
of ongoing work: 
\begin{itemize}

\item \textit{Commerce, Brand Management, Customer Relationship Management, and FinTech (Finance and Technology)}: Sentiment analysis of blogs, tweets, and Facebook posts is already
widely used to shape brand image, track customer response, and in developing automatic dialogue systems for handling customer queries and complaints 
\cite{ren2012linguistic,yen2014emotional}.\\[-19pt] 

\item \textit{Art}: Art and emotions are known to have a strong connection---art often tends to be evocative. Thus it is not surprising to see the growing use of emotion resources in art. \cite{davis14} developed a system, \textit{TransProse}, that generates music that captures the emotions in a piece of literature. The system uses the NRC Emotion Lexicon.
A symphony orchestra performed music composed using TransProse and a human composer at the Louvre museum in Paris, September 20, 2016.
Wishing Wall, an interactive art installation that visualizes wishes using the NRC Emotion lexicon was displayed in Tekniska Museet, Stockholm, Sweden (Oct 2014--Aug 2015), Onassis Cultural Centre, Athens (Oct 2015--Jan 2016), and Zorlu Centre in Istanbul (Feb 12th--June16).
WikiArt Emotions that annotated emotions evoked by art \cite{mohammad-kiritchenko-2018-wikiart}.\\[-19pt]

\item \textit{Education}: Automatic tutoring and student evaluation systems detect emotions in responses to determine correctness of responses and also
to determine emotional state of the participant \cite{li2014study,suero2014emotion}. It has been shown that
learning improves when the student is in a happy and calm state as opposed to anxious or frustrated \cite{dogan2012emotion}.\\[-19pt]

\item\textit{ Tracking The Flow of Emotions in Social Media}: Besides work in 
brand management and public health, as discussed already, some recent work attempts to better understand how emotional information {\it spreads} in a social network, 
for instance to improve disaster management 
\cite{kramer2012spread,vo2013twitter}.\\[-19pt]


\item \textit{Literary Analysis and Digital Humanities}: There is growing interest in using automatic natural language processing techniques to analyze
large collections of literary texts. Specifically with respect to emotions, there is work on tracking the flow of emotions
in novels, plays, and movie scripts, detecting patterns of sentiment common to large collections of texts, and tracking emotions of
plot characters 
\cite{mohammad:2011:LaTeCH-2011,Mohammad2012730,hartner2013lingering}. 
There is also work in generating music that captures the emotions in text \cite{davis14}.\\[-19pt]

\item \textit{Personality Traits}: Systematic patterns in how people express emotions are key indicators
of personality traits such as extroversion and narcissism. Thus many automatic systems that determine personality traits 
from written text rely on automatic detection of emotions
\cite{grijalva2014gender,ICWSM136138,MohammadK13}.\\[-19pt] 

\item \textit{Understanding Social Groups}: People of various genders, races, and social groups have often encountered different and unequal social conditions. Several studies
have analyzed the differences in emotions in language used by different social groups and in language mentioning people from different social groups 
\cite{mohammad-yang:2011:WASSA2011,grijalva2014gender,montero2014investigating}.\\[-19pt]

\item \textit{Politics}: There is tremendous interest in tracking public sentiment, especially in social media, towards politicians, electoral issues, as well as national and international events.
Some studies have shown that the more partisan electorate tend to tweet more, as do members from
minority groups \cite{lassen2011twitter}.  
There is work on identifying contentious issues \cite{Maynard11} and on detecting voter polarization \cite{Conover11}.  
Tweet streams have been shown to help identify current
public opinion towards the candidates in an election (nowcasting)
\cite{Golbeck11,Mohammad2014elec}.  
Some research has also shown
the predictive power of analyzing electoral tweets to determine the number of votes a candidate will
get (forecasting) \cite{Tumasjan10,lampos2013user}. 
However, other research expresses
skepticism at the extent to which forecasting is possible \cite{Avello12}.\\[-19pt]

\item \textit{Public Health and Psychology}: Automatic methods for detecting emotions are useful in detecting depression 
\cite{pennebaker2003psychological,cherry2012binary}, 
identifying cases of cyber-bullying \cite{chen2012detecting}, 
predicting health attributes at community level \cite{johnsen2014language,eichstaedt2015psychological},
gender differences in how we perceive connotative word meaning \cite{vad-acl2018},
and tracking well-being \cite{ICWSM136138,paul2011you}.

There is also interest in robotic assistants and physio-therapists for the elderly, the disabled, and the sick---robots
that are sensitive to the emotional state of the patient.\\[-19pt]


\item \textit{Visualizing Emotions}: A number of applications listed above benefit from good visualizations of emotions in text(s).
Particularly useful is the feature of interactivity. If users are able to select particular aspects such as an entity, emotion, or time-frame of interest,
and the system responds to show information relevant to the selection in more detail, then the visualization enables improved user-driven exploration of the data.
See  
\citeA{quan2014visualizing,Mohammad2012730,liu2003visualizing} 
for work on visualization of emotions in text.\\[-16pt]
\end{itemize}
\noindent As automatic methods to detect various affect categories become more accurate,
their use in natural language applications will likely become even more ubiquitous.

\section{Ethics and Fairness}

Recent advances in machine learning have meant that computer-aided systems are becoming more human-like in their predictions. This also means that they perpetuate human biases. Some learned biases may be beneficial for the downstream application. Other biases can be inappropriate and result in negative experiences for some users. Examples include, loan eligibility and crime recidivism systems that negatively assess people belonging to a certain area code (which may disproportionately impact people of a certain race) \cite{chouldechova2017fair} and resum\'e sorting systems that believe that men are more qualified to be programmers than women \cite{bolukbasi2016man}. Similarly, sentiment and emotion analysis systems can also perpetuate and accentuate inappropriate human biases. In fact, recent work has shown that a large majority of sentiment and emotion analysis machine learning systems consistently give different emotionality scores to sentences mentioning different races and genders (Kiritchenko and Mohammad, 2018). Often these biases are across stereotypical lines, for example, marking near-identical sentences that mention African Americans names as being more angry than sentences that mention European American names, and sentences that mention women as being more emotional (more happy / more sad) than sentences that mention men. 
\citeA{thelwall2018gender} found that automatic sentiment analysis systems are somewhat less accurate on customer reviews written by men compared to those written by women. 
\citeA{diaz2018addressing} examined several sentiment resources, including word embeddings, to find that several old-age-related terms (and their mentions in text) are marked as having negative sentiment whereas terms associated with the young (and their mentions in text) are often evaluated as being positive. Thus sentiment analysis in the real-world has potential negative implications for the elderly. All of this has brought greater attention on how we create and use emotion resources (training data, lexicons, algorithms, etc.) and their role in creating fair emotion systems. For example, see \citeA{mohammad2020practical} for a description of various ethical considerations in the use of word--emotion association lexicons.

Sentiment analysis, like other natural language technologies, can be a \textit{great enabler}---allowing one to capitalize on substantial, previously unattainable, opportunities. However, the opportunities can also be used by bad actors for committing maleficence. For example, imagine a world where sentiment analysis of social media posts can easily lead to determining one's position on a wide variety of issues, likes and dislikes, personality traits, as well as the emotional state when most pliable for persuasion (say, to purchase items online or to support a political agenda). Suddenly, we (both as individuals and whole populations) are open for manipulation, deception, and indoctrination. Unfortunately, that world is already upon us. To name just two such instances from the recent past: Facebook was reported to have told advertisers that it can identify teenagers who ``feel stressed”, ``defeated”, ``overwhelmed”, ``anxious”, ``nervous”, ``stupid”, ``silly”, ``useless” and a ``failure”.\footnote{https://www.theguardian.com/technology/2017/may/01/facebook-advertising-data-insecure-teens} Cambridge Analytica was accused to illegally extracting profiles of millions of Facebook users and leverage it to sway public opinion in the 2016 US elections \cite{cadwalladr2018revealed}. 
Thus, despite the many benefits of sentiment analysis (such as those stated in the previous section), both  the researcher and the lay person have to be on guard for the perils it will inevitably germinate.


\section{Summary and Future Directions}
This chapter summarized the diverse landscape of problems and applications associated with
automatic sentiment analysis. We outlined key challenges for automatic systems, as well as
the algorithms, features, and datasets used in sentiment analysis. We described several
manual and automatic approaches to creating valence- and emotion-association lexicons.
We also described work on sentence-level sentiment analysis. 
We discussed work preliminary work on sentiment modifiers such as negators and modals, 
work on detecting sentiment in figurative and metaphoric language, as well as 
cross-lingual sentiment analysis---these are areas where we expect to see significantly
more work in the near future. Other promising  areas of future work include: understanding the relationships
between emotions, multimodal affect analysis (involving not just text but also speech, vision, physiological sensors, etc),
and applying emotion detection to new applications.

Natural Language Processing (NLP), the broader field that encompasses sentiment analysis, has always been
an interdisciplinary field with strong influences from Computer Science, Linguistics, and Information Sciences. However, over the last decade, NLP and Sentiment Analysis have made marked inroads into other fields of study
such as psychology, digital humanities, history, art, and social sciences---sometimes attracting controversy. (For example, see the articles on the so called Digital Humanities Wars.\footnote{https://www.chronicle.com/article/The-Digital-Humanities-Debacle/245986})
A common principle in these advances is that NLP allows for asking old and new questions in these fields
by examining massive amounts of text. Over the coming decade, we expect more of this, with the traditional boundaries between NLP and other fields becoming even more blurry, along with a healthy cross-pollination of ideas. Finally, we also expect to see considerable interest in the ethics, fairness, and bias of sentiment analysis systems, as well as the use of sentiment analysis to help identify explicit and implicit biases of people.


\bibliography{skinny-ref}
\bibliographystyle{theapa}

\end{document}